%% file: main.tex
\documentclass[sigconf]{acmart}

\AtBeginDocument{%
  }

\copyrightyear{2026}
\acmYear{2026}
\setcopyright{cc}
\setcctype{by}
\acmConference[MM '26]{Proceedings of the 34th ACM International Conference on Multimedia}{November 10--14, 2026}{Rio de Janeiro, Brazil}
\acmBooktitle{Proceedings of the 34th ACM International Conference on Multimedia (MM '26), November 10--14, 2026, Rio de Janeiro, Brazil}
\acmDOI{10.1145/3767308.3835123}
\acmISBN{979-8-4007-2213-4/2026/11}

\input{preamble}

\begin{document}

\title{See More, Detect Less? Taming Information Leakage in Multi-View Anomaly Detection}

\author{Shang-Fu Chen}
\orcid{0009-0004-9371-0892}
\affiliation{%
  \institution{National Taiwan University}
  \city{Taipei}
  \country{Taiwan}
}
\email{chenshangfu@cmlab.csie.ntu.edu.tw}

\author{Kuan-Chuan Peng}
\orcid{0000-0002-2682-9912}
\affiliation{%
  \institution{Mitsubishi Electric Research Laboratories (MERL)}
  \city{Cambridge}
  \state{MA}
  \country{USA}
}
\email{kpeng@merl.com}

\author{Jhih-Ciang Wu}
\orcid{0000-0003-4071-3980}
\affiliation{%
  \institution{National Taiwan Normal University}
  \city{Taipei}
  \country{Taiwan}
}
\email{jcwu@csie.ntnu.edu.tw}

\author{Wen-Huang Cheng}
\orcid{0000-0002-4662-7875}
\affiliation{%
  \institution{National Taiwan University}
  \city{Taipei}
  \country{Taiwan}
}
\affiliation{
\institution{VinUniversity}
\city{Hanoi}
\country{Vietnam}
}
\email{wenhuang@csie.ntu.edu.tw}

\author{Kai-Lung Hua}
\orcid{0000-0002-7735-243X}
\affiliation{%
  \institution{Microsoft Taiwan Corporation, National Taiwan University of Science and Technology}
  \city{Taipei}
  \country{Taiwan}}
\email{kai.hua@microsoft.com}

\renewcommand{\shortauthors}{Shang-Fu Chen, Kuan-Chuan Peng, Jhih-Ciang Wu, Wen-Huang Cheng, \& Kai-Lung Hua}

\input{0_abstract}






\maketitle

\input{1_intro}
\input{2_related-work}

\input{3_method}
\input{4_experiments}

\input{5_conclusion}


\bibliographystyle{ACM-Reference-Format}
\bibliography{main}



\end{document}

%% file: 0_abstract.tex
\begin{abstract}
In multi-view anomaly detection, more cross-view information can actually hurt. When multiple inspection views are naively fused in a reconstruction-based pipeline, normal cues from intact views propagate to the decoder, which faithfully reconstructs anomalous regions, collapsing the reconstruction gap the detector depends on. We call this failure mode \emph{cross-view information leakage} and show that effective multi-view fusion must explicitly restrict the information reaching the decoder. Building on this insight, we present \ours (Global-Local Attention Driven framework), the first framework combining vision foundation model features with local and global cross-view fusion for multi-view anomaly detection. The Multi-view Merging Attention (MMA) module performs local cross-view fusion at linear complexity with learnable view importance weighting and token-wise gating, letting each view selectively incorporate fine-grained evidence from other views at $\mathcal{O}(N)$ cost. The Object-Guided Attention (OGA) module captures global context by aggregating class tokens from all views into a single object-level representation and broadcasting it back to patch tokens via temperature-scaled sigmoid gating, replacing the original patch representations rather than adding a residual to preserve the reconstruction gap. Experiments on Real-IAD and MANTA-Tiny show that \ours outperforms state-of-the-art methods across sample-, image-, and pixel-level metrics, confirming that principled information restriction is key to multi-view anomaly reasoning.
\end{abstract}

%% file: 1_intro.tex
\section{Introduction}
\label{sec:intro}

\wu{\SC{Industrial inspection systems~~\cite{wang2024real,fan2024manta}} often capture each object from multiple camera angles to achieve complete surface coverage. This acquisition process naturally produces a multi-view visual signal, in which different viewpoints provide complementary evidence about the object's geometry and appearance. Multi-view anomaly detection (AD) in such a setting is of great practical importance for industrial quality control. However, \SC{most existing approaches~\cite{zhang2025exploring,guo2025dinomaly,gu2025univad,zhang2025wave}} still treat each view as an independent sample, largely ignoring the cross-view relationships inherent in the data. At the same time, recent progress in single-view anomaly detection~\cite{deng2024simclip,wang2025uniad,fang2025af} has been substantial, particularly with reconstruction-based frameworks~\cite{guo2025dinomaly,zhang2025exploring} built upon Vision Transformer (ViT) architectures~\cite{deng2022anomaly,tien2023revisiting,you2022unified,he2024mambaad,zhang2025exploring}. These methods typically freeze a pre-trained encoder and train a lightweight decoder to reconstruct only normal patterns, enabling anomalies to be identified through reconstruction errors. While effective for single-view inspection, this paradigm does not fully suit the multi-view setting, where cross-view consistency is essential for resolving ambiguities in individual viewpoints. In practice, a defect that is difficult to recognize in one view may become evident in another. Motivated by this, recent benchmarks such as Real-IAD~\cite{wang2024real} and MANTA~\cite{fan2024manta} have brought multi-view anomaly detection to the forefront. Yet existing methods~\cite{he2024learning,mao2025unveiling} remain limited, as they either fail to model global object-level interactions across views or introduce substantial computational overhead for high-dimensional feature fusion.}

\input{fig_teaser}

\wu{\SC{Vision foundation models~\cite{oquab2023dinov2,deng2024simclip,damm2025anomalydino,ma2025aaclip,fang2025af}} have shown strong performance in single-view AD by providing rich, pre-trained representations. Within \SC{reconstruction-based frameworks~\cite{deng2022anomaly,guo2025dinomaly,zhang2025exploring,liu2025unlocking}}, a frozen ViT encoder extracts these features, while a lightweight decoder learns to reproduce normal patterns, revealing anomalies through reconstruction discrepancies. However, extending these powerful representations to effective multi-view reasoning remains unexplored. To this end, we propose the Global-Local Attention Driven (\ours) framework, the first approach that integrates vision foundation model features with both global and local cross-view fusion for multi-view AD. \ours encodes each view independently using a shared frozen encoder and subsequently fuses the resulting token sequences across views before decoding. This design preserves the strengths of single-view representations while enabling structured inter-view interactions.
A central challenge in this setting is computational efficiency, as naive cross-attention over all tokens and views incurs quadratic complexity in both the token count and the number of views. To address this, we design a global-local fusion pipeline composed of two complementary modules. The \textbf{Multi-view Merging Attention (MMA)} module performs local cross-view fusion, reducing the cost from $\mathcal{O}(N^2)$ to $\mathcal{O}(N)$, where N is the number of tokens per view. It further incorporates a learnable view-importance matrix to adaptively weight different viewpoints, along with a token-wise gating mechanism to regulate the contribution of fused features. The \textbf{Object-Guided Attention (OGA)} module captures global object-level context by aggregating class tokens from all views into a unified object representation via a learned projection. This global token is then propagated back to patch tokens through cross-attention with temperature-scaled sigmoid gating, enabling each view to be conditioned on holistic object-level information.
}

\wu{\SC{In standard multi-view learning~\cite{yu2024review,wang2025seqmvrl,dong2025enhanced,xu2025robust}}, unconstrained cross-view information flow is typically considered beneficial. However, we find that reconstruction-based AD breaks this intuition. When multiple views are naively fused, normal cues from intact viewpoints can propagate to the decoder, enabling it to faithfully reconstruct even anomalous regions. This phenomenon, which we define as \emph{cross-view information leakage}, effectively collapses the ``reconstruction gap'' upon which AD relies. 
\SC{As shown in Figure~\ref{fig:teaser}, na\"ive fusion obscures the anomaly signal and collapses the normal-abnormal reconstruction gap, while our information restricted design restores it.}
Our analysis reveals a critical insight: effective multi-view fusion in this setting demands explicit information restriction rather than unconstrained sharing. This insight directly informs the design of the OGA module. \SC{Unlike conventional fusion~\cite{vaswani2017attention}} that adds features as a residual, OGA replaces original patch representations with globally conditioned outputs, thereby limiting the decoder's access to view-specific normal details that might otherwise facilitate ``leaked'' reconstructions. Furthermore, we employ temperature-scaled sigmoid gating instead of standard softmax normalization to avoid forcing competition among normal tokens, using a sharpened temperature to concentrate attention on a sparse set of relevant positions. These deliberate constraints ensure that the decoder remains conditioned on holistic object context rather than local anomalous cues, thereby preserving the reconstruction gap.}

We evaluate \ours on the MANTA~\cite{fan2024manta} and Real-IAD~\cite{wang2024real} benchmarks, showing consistent improvements at the sample, image, and pixel levels over both single-view baselines and the state-of-the-art methods. Our main contributions are summarized as follows:
\begin{itemize}
    \item We propose \ours, a global-local multi-view anomaly detection framework built on vision foundation models that integrates object-level global context with token-level local correspondences to achieve state-of-the-art performance.
    \item We introduce the Multi-view Merging Attention (MMA) module, a linear-complexity cross-view mechanism that enables scalable local fusion through learnable view-importance weighting and adaptive token-wise gating.
    \item We design the Object-Guided Attention (OGA) module, which constructs a holistic object representation and employs an information-restricted broadcast mechanism to prevent the collapse of the reconstruction gap.
    \item We identify and analyze cross-view information leakage, a fundamental challenge in multi-view reconstruction, and demonstrate that strategic information restriction is essential for robust defect localization.
\end{itemize}

%% file: fig_teaser.tex
\begin{figure}[ht!]
    \centering
    \includegraphics[width=.95\linewidth]{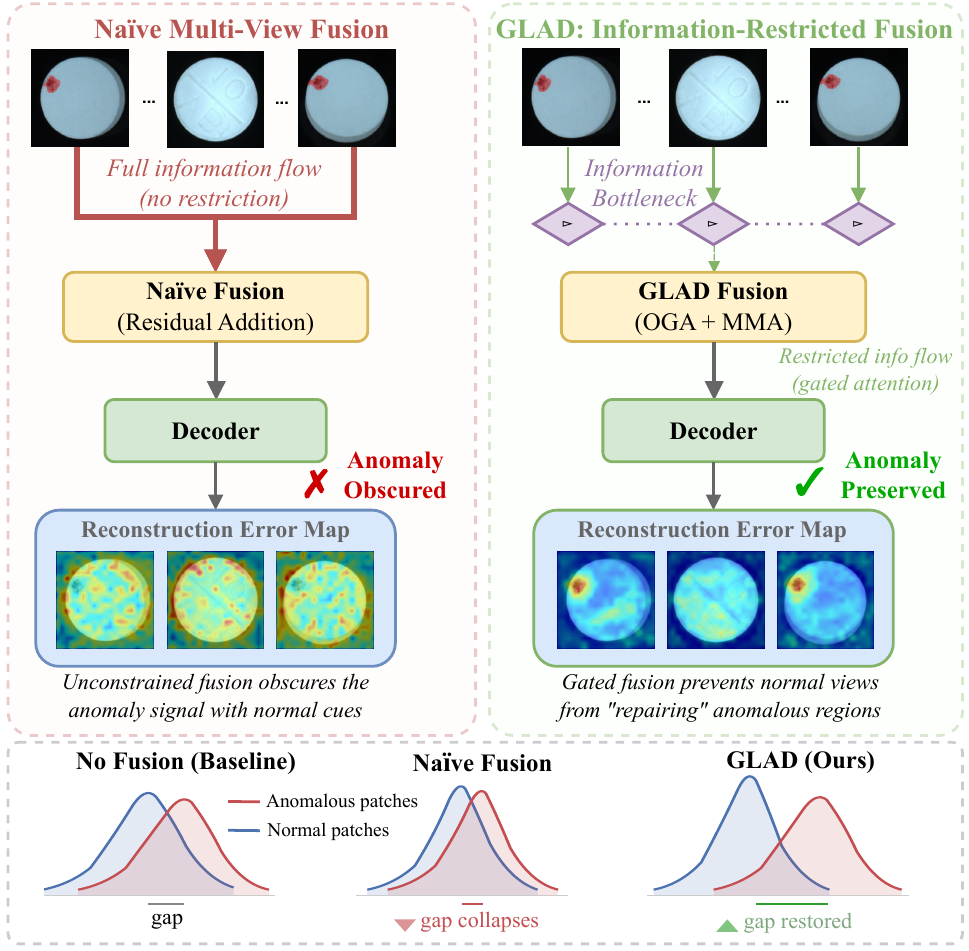}
    \Description[<short description>]{<long description>}
    \vspace{-1em}
    \caption{
    Cross-view information leakage in multi-view anomaly detection. Unconstrained fusion obscures anomalies by collapsing the normal–abnormal reconstruction gap.
    \SC{\ours restores this gap through information-restricted fusion.}
    }
    \vspace{-1em}
    \label{fig:teaser}
\end{figure}

%% file: 2_related-work.tex
\section{Related Work}
\label{sec:related-work}

\noindent \textbf{Anomaly Detection (AD).}
Unsupervised AD methods identify defects using only normal training samples.
Table~\ref{tab:related_works} categorizes existing single-view AD methods into three groups according to their modeling scope and backbone choice: separate methods that train one model per class ($\gG_0$), unified methods built on conventional backbones ($\gG_1$), and unified methods that leverage vision foundation models ($\gG_2$).
\noindent Category~$\gG_2$~\cite{zhang2025exploring,guo2025dinomaly} adopt frozen foundation models as encoders to obtain rich, generalizable features.
\SC{ViTAD~\cite{zhang2025exploring} performs plain ViT feature regression on DINO~\cite{oquab2023dinov2} features, and Dinomaly~\cite{guo2025dinomaly} reconstructs foundation-model features with a noisy bottleneck.}
These approaches achieve strong single-view results but still treat each view independently, discarding valuable cross-view correlations.
\SC{\noindent Category~$\gG_1$~\cite{you2022unified,he2024diffusion,kashiani2025roads,wei2025uninet,he2024mambaad,zhang2025wave} handle all object classes in a single model.}
\SC{DiffusionAD~\cite{he2024diffusion} formulates AD as a denoising diffusion process, ROADS~\cite{kashiani2025roads} leverages reverse distillation with an online learning strategy, and UniNet~\cite{wei2025uninet} fuses multi-layer features within a unified transformer.
MambaAD~\cite{he2024mambaad} and WaveMambaAD~\cite{zhang2025wave} explore state-space models as decoders for efficient long-range modeling, while retaining pre-trained CNN encoders for feature extraction.}
Although these methods reduce deployment overhead, their conventional backbones struggle with the diverse appearances in multi-view settings.
\noindent Category~$\gG_0$~\cite{li2021cutpaste,liu2023simplenet,hu2024anomalydiffusion,roth2022towards,defard2021padim,deng2022anomaly,zavrtanik2021draem,chen2025filter} dedicate one model to each object class, yielding competitive accuracy at the cost of scalability.
Synthesis-based approaches~\cite{li2021cutpaste,liu2023simplenet,hu2024anomalydiffusion} generate artificial defects but face a synthetic-to-real gap that widens across viewpoints.
Embedding-based approaches~\cite{roth2022towards,defard2021padim} compare test features against a stored normal distribution, but lack cross-view spatial correspondence.
Reconstruction-based approaches~\cite{deng2022anomaly,zavrtanik2021draem,chen2025filter} learn to reproduce normal patterns and flag deviations, but without cross-view constraints can miss anomalies visible only from certain angles.
In contrast, \ours builds on foundation-model features while introducing multi-view fusion through linear-complexity local cross-view attention (MMA) and information-restricted global object-level aggregation (OGA), where cross-view information flow is controlled to preserve the reconstruction gap that AD relies on.

\noindent \textbf{Multi-view AD.}
Multi-view AD is widely adopted in industrial quality control, yet existing single-view methods discard cross-view consistency cues and therefore cannot exploit the complementary information captured from different viewpoints.
MVAD~\cite{he2024learning} is among the first to jointly process multiple views via patch-level cross-attention with an adaptive view selection mechanism, but its top-$k$ token interactions remain memory-intensive, and it lacks a global, object-level representation, so sample-level decisions rely solely on aggregating local patch scores.
IDIF~\cite{mao2025unveiling} instead decouples intra-view features into a shared voxel space for geometric cross-view alignment, but its class-specific design prevents scaling to unified multi-class deployment.
Recent benchmarks such as Real-IAD~\cite{wang2024real} and MANTA~\cite{fan2024manta} provide multi-view evaluation protocols that further expose these limitations, revealing a clear gap between single-view performance and the demands of holistic multi-view reasoning.
As summarized in Table~\ref{tab:related_works}, \ours is the first method that simultaneously satisfies all five desirable properties: it adopts a vision foundation model to get rich representation, operates in the unified setting, accepts multi-view input, and performs both local pixel-level fusion via MMA and global object-level fusion via OGA.

\input{tab_related-work}

\input{fig_method}

\noindent \textbf{Global-to-Local Attention.}
Using a global or class-level token to refine local features via cross-attention has proven effective across vision tasks.
\SC{CaiT~\cite{touvron2021cait} separates class-attention layers from patch self-attention, and CrossViT~\cite{chen2021crossvit} performs multi-scale fusion via class token cross-attention.}
This idea is also used in dense prediction, where MCTformer~\cite{xu2022mctformer} and CTI~\cite{yoon2024cti} use class tokens to generate class-specific attention maps for weakly supervised segmentation.
More broadly, learned query tokens~\cite{carion2020detr,li2023blip2,jaegle2021perceiver} that cross-attend over input features are a common design pattern.
However, these methods operate on single images and do not address how to construct a meaningful global token that captures information shared across multiple views of the same object.
\SC{Our OGA module fills this gap by aggregating CLS tokens from all views into a unified global object token and cross-attending it over each view's patch tokens with temperature-scaled sigmoid gating. Unlike standard cross-attention, OGA replaces rather than adding a residual to prevent cross-view information leakage.}
Beyond global fusion, \ours further pairs OGA with a local token-level cross-view mechanism (MMA), enabling the model to capture both object-level consistency and fine-grained spatial correspondences across views.

%% file: tab_related-work.tex


\begin{table}[!t]
\caption{Characteristic comparison between \ourso and prior anomaly detection (AD) works ($\gG_0$: \cite{li2021cutpaste,liu2023simplenet,hu2024anomalydiffusion,roth2022towards,defard2021padim,deng2022anomaly,zavrtanik2021draem,chen2025filter}, $\gG_1$: \cite{you2022unified,he2024diffusion,kashiani2025roads,wei2025uninet,he2024mambaad,zhang2025wave}, $\gG_2$: \cite{zhang2025exploring,guo2025dinomaly}).}
\vspace{-1em}
\setlength{\tabcolsep}{4pt}
\resizebox{\columnwidth}{!}{
\begin{tabular}{rc@{\hspace{3mm}}c@{\hspace{3mm}}c@{\hspace{3mm}}c@{\hspace{3mm}}c>{\columncolor{OursColor}}c}
\toprule
\multirow{2.5}{*}{\textbf{\shortstack{AD Method Conditions}}} & \multicolumn{6}{c}{\textbf{AD Categories}} \\
\cmidrule(l){2-7}
 & $\gG_0$ & $\gG_1$ & $\gG_2$ & IDIF~\cite{mao2025unveiling} & MVAD~\cite{he2024learning} & \ours \\
\midrule
Single model for all classes & \ccross & \ccheck & \ccheck & \ccross & \ccheck  & \ccheck \\
Multi-view input & \ccross & \ccross & \ccross & \ccheck & \ccheck & \ccheck \\
Vision foundation model & \ccross & \ccross & \ccheck & \ccross & \ccross & \ccheck \\
Token-Level cross-view fusion & \ccross & \ccross & \ccross & \ccheck & \ccheck & \ccheck \\
Object-Level cross-view fusion & \ccross & \ccross & \ccross & \ccross & \ccross & \ccheck \\
\bottomrule
\end{tabular}
}
\label{tab:related_works}
\vspace{-1em}
\end{table}

%% file: fig_method.tex
\begin{figure*}[ht]
    \centering
    \includegraphics[width=.95\textwidth]{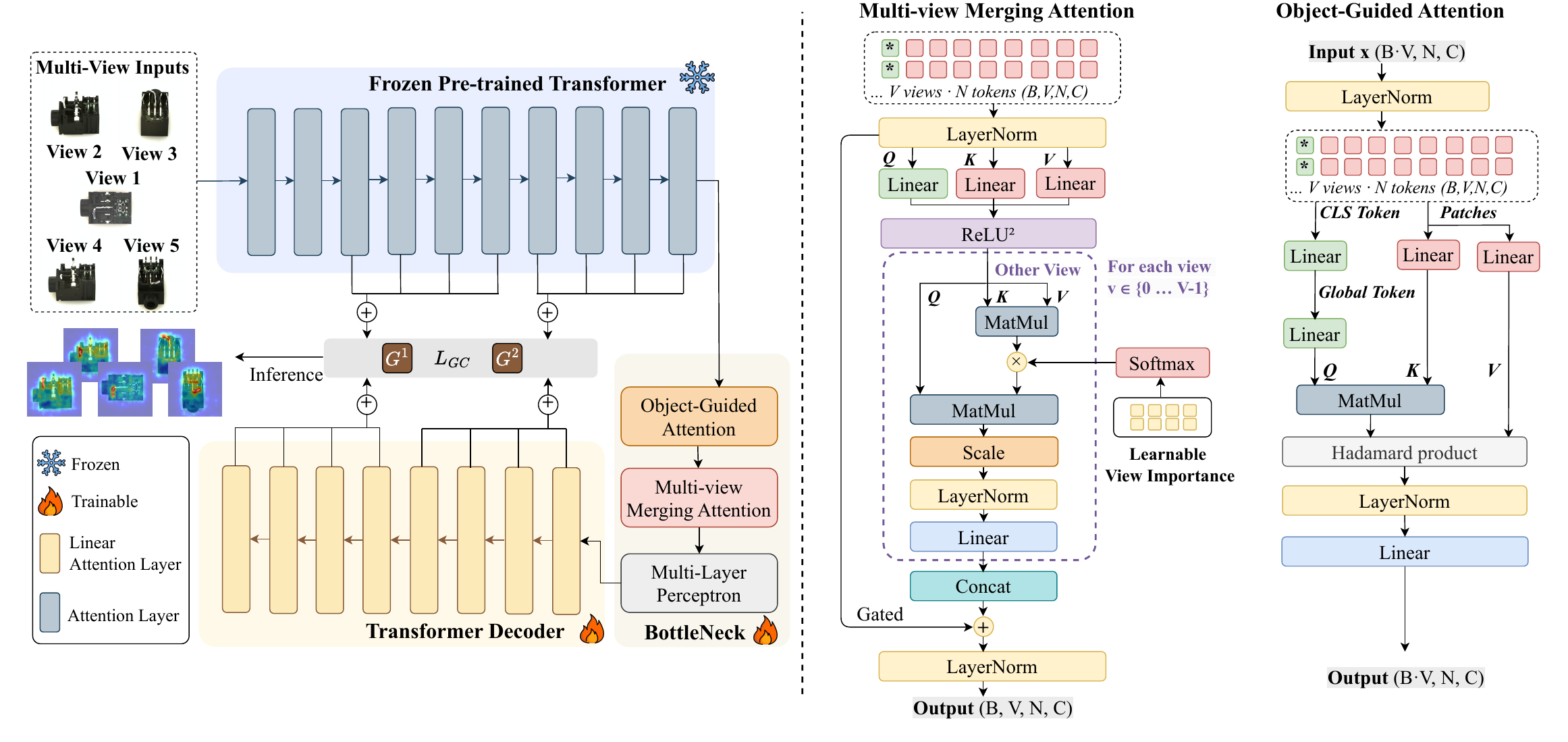}
    \Description[<short description>]{<long description>}
    \vspace{-1.5em}
    \caption{
    Overview of \ours. Multi-view images are encoded by a frozen pre-trained Transformer. The resulting tokens pass through the Object-Guided Attention (OGA) module for global object-level fusion and the Multi-view Merging Attention (MMA) module for local token-level cross-view fusion, before entering the MLP bottleneck and Transformer decoder for reconstruction. Right panels detail the MMA (middle) and OGA (right) architectures.
    }
    \vspace{-1em}
    \label{fig:architecture}
\end{figure*}

%% file: 3_method.tex
\section{Our Proposed Method -- \ours}
\label{sec:method}

\subsection{Overview}
\label{subsec:overview}


\noindent \textbf{Problem Setup.}
Given an object captured from $V$ viewpoints, the input consists of $V$ images $\{\mathbf{x}_v\}_{v=1}^{V}$ with $\mathbf{x}_v \in \mathbb{R}^{H \times W \times 3}$.
The goal is to determine whether the object is anomalous at the sample level and, for each view, to produce a pixel-level anomaly map that localizes defective regions.
Only normal images are available during training.
As shown in Figure~\ref{fig:architecture}, \ours follows the encoder-bottleneck-decoder reconstruction paradigm~\cite{guo2025dinomaly,zhang2025exploring}, where a shared, frozen self-supervised Vision Transformer encodes each view independently and a lightweight Transformer decoder is trained to reconstruct the encoder features.
For view $v$, the encoder produces a class token $\mathbf{z}_v^{\text{cls}} \in \mathbb{R}^{D}$ that summarizes global content and a set of patch tokens $\mathbf{Z}_v \in \mathbb{R}^{N \times D}$ that encodes local spatial information, where $N$ is the number of patches and $D$ is the feature dimension.
We extract features from multiple intermediate encoder layers and average them to form a single token sequence per view.
The anomaly map for each view is computed as the cosine distance between the encoder output and the decoder reconstruction at every spatial position, and the sample-level score aggregates evidence across all $V$ views.
Following~\cite{guo2025dinomaly}, we compute a global cosine similarity loss $L_{GC}$ between the encoder and decoder feature maps across all spatial patches. To focus optimization on informative regions, we apply a hard-mining strategy during backpropagation, which patches are ranked by reconstruction error and partitioned into an easy group $G^1$ (top $p\%$ with lowest error) and a hard group $G^2$ (remaining $(1{-}p)\%$). The gradients of $G^1$ are scaled by a near-zero factor, so that parameter updates are predominantly driven by the hard-to-reconstruct patches in $G^2$. 


\noindent \textbf{Multi-view Fusion.}
Existing single-view methods pass the encoder tokens directly through an MLP bottleneck and into the decoder, treating each view in isolation.
\ours instead inserts a multi-view fusion stage between the encoder and the bottleneck, enabling cross-view reasoning before reconstruction (Figure~\ref{fig:architecture}).
This stage comprises two complementary modules.
First, Object-Guided Attention (OGA, Sec.~\ref{subsec:OGA}) aggregates the class tokens $\{\mathbf{z}_v^{\text{cls}}\}_{v=1}^{V}$ from all views into a single global object token via a learned projection, then cross-attends this token over patch tokens across all views and broadcasts the result back as a gated residual, injecting holistic object-level context into every view.
Second, Multi-view Merging Attention (MMA, Sec.~\ref{subsec:MMA}) performs token-level cross-view fusion with linear complexity, where each view attends to patch tokens from the remaining views using an ReLU\textsuperscript{2} kernel, weighted by a learnable view importance matrix and modulated by a token-wise gate.
Combining global (OGA) and local (MMA) fusion lets the decoder reconstruct features informed by both object-level and fine-grained cross-view cues.


\subsection{Object-Guided Attention (OGA)}
\label{subsec:OGA}

\SC{
A patch token knows what it sees, but not what it is part of. When multiple views capture the same object, each patch encodes a local surface fragment without awareness of the object's overall identity or how the other views perceive it. Before fusing token-level details across views, \ours therefore first establishes a holistic understanding of the object. OGA fills this gap by constructing a single global object token from all views and using it to modulate every patch token through cross-attention, giving each local feature a sense of the whole object before reconstruction begins.
}

\noindent \textbf{Global Object Token.}
To obtain a compact, object-level representation, OGA concatenates the class tokens from all $V$ views and projects them through a learned linear layer:
\begin{equation}
\label{eq:oga_global}
\mathbf{g} = \mathbf{W}_g \,[\mathbf{z}_1^{\text{cls}};\, \mathbf{z}_2^{\text{cls}};\, \ldots;\, \mathbf{z}_V^{\text{cls}}]\,,
\end{equation}
where $\mathbf{W}_g \in \mathbb{R}^{D \times VD}$ is a learnable projection and $[\,\cdot\,;\,\cdot\,]$ denotes concatenation.
Each class token $\mathbf{z}_v^{\text{cls}}$ encodes the global semantics of a single view; by fusing all of them, $\mathbf{g} \in \mathbb{R}^{D}$ captures an object-level summary that is aware of all viewpoints simultaneously.

\noindent \textbf{Cross-Attention over Multi-view Patches.}
The global object token is used as a query to cross-attend over the patch tokens from all views.
Specifically, all $V$ sets of patch tokens are concatenated into a single sequence $\mathbf{P} = [\mathbf{Z}_1;\, \mathbf{Z}_2;\, \ldots;\, \mathbf{Z}_V] \in \mathbb{R}^{VN \times D}$, and multi-head projections are applied:
\begin{equation}
\label{eq:oga_qkv}
\mathbf{q} = \mathbf{g}\,\mathbf{W}^Q, \quad
\mathbf{K} = \mathbf{P}\,\mathbf{W}^K, \quad
\mathbf{V} = \mathbf{P}\,\mathbf{W}^V\,,
\end{equation}
where the query $\mathbf{q} \in \mathbb{R}^{N_h \times 1 \times d}$ and the keys and values $\mathbf{K}, \mathbf{V} \in \mathbb{R}^{N_h \times VN \times d}$ are reshaped into $N_h$ heads of dimension $d = D / N_h$.
The attention weights are computed with a temperature-scaled sigmoid activation:
\begin{equation}
\label{eq:oga_attn}
\mathbf{A} = \sigma\!\left(\frac{\mathbf{q}\,\mathbf{K}^\top}{\tau\sqrt{d}}\right)\,,
\end{equation}
where $\sigma$ denotes the sigmoid function and $\tau$ is a temperature hyperparameter (set to $0.7$).
Each entry $A_i \in [0,1]$ acts as an independent gate that controls how much information from the $i$-th patch flows into the output.
The per-patch output is then obtained by element-wise gating rather than a weighted sum:
\begin{equation}
\label{eq:oga_output}
\hat{\mathbf{z}}_i = VN \cdot A_i \cdot \mathbf{v}_i\,,
\end{equation}
where $\mathbf{v}_i$ is the value vector of the $i$-th patch and the scaling factor $VN$ compensates the reduced magnitude due to sigmoid gating (unlike softmax, does not guarantee that the weights sum to one).

\noindent \textbf{Information-Restricted Broadcast.}
In reconstruction-based AD, the model relies on a gap between reconstruction errors for normal and anomalous inputs.
A central challenge in designing OGA is that the global token should amplify the difference between normal and abnormal regions rather than smooth everything toward a normal template.
If the cross-attention output is naively added as a residual, the global token effectively produces a normal template, because the vast majority of training patches are normal.
At test time, this template would push anomalous regions closer to the normal distribution, collapsing the reconstruction gap that the detector depends on.
We refer to this failure mode as \emph{cross-view information leakage} and analyze it empirically in Sec.~\ref{sec:experiments}.
\SC{
To prevent leakage, OGA operates in a \emph{token replacement} mode, where the cross-attention output directly replaces the original patch tokens instead of being added as a residual.
}
By discarding the original features entirely, the model avoids accumulating a normal template that would dilute the anomaly signal:
\begin{equation}
\label{eq:oga_replace}
\hat{\mathbf{Z}}_v = \text{LN}\!\left(\text{OGA}(\mathbf{Z}_v)\right)\,,
\end{equation}
where LN denotes the layer normalization~\cite{ba2016layer}. The attention mechanism itself is also designed to sharpen the normal-abnormal distinction.
OGA uses sigmoid rather than softmax in Eq.~\eqref{eq:oga_attn}, because softmax normalizes across all patches, forcing normal patches to compete with each other for attention weight and reducing the mechanism to a normal-vs-normal comparison that is insensitive to anomalies.
Sigmoid gates each patch independently, allowing abnormal patches to receive distinctly different activation levels without being diluted by the normal majority.
A temperature $\tau = 0.7 < 1$ further sharpens the distribution, widening the gap between the activation levels assigned to normal and abnormal patches.
Finally, the global object token $\mathbf{g}$ is \emph{not} added back to the per-view class tokens after cross-attention, preserving the single-view nature of the class token for downstream anomaly scoring.
These choices ensure that OGA injects global object context into each view's patch tokens in a way that amplifies the normal-abnormal distinction, rather than blurring it, while preserving the reconstruction bottleneck that is essential for AD.

\subsection{Multi-view Merging Attention (MMA)}
\label{subsec:MMA}

\SC{
While OGA provides each patch token with global object-level context, it does not enable direct communication between individual patches across views.
Consider a scratch on an object that is clearly visible in views 2 and 4 but occluded in views 1, 3, and 5. OGA can inform all patches that the object has a certain global profile, but it cannot propagate the fine-grained evidence of the scratch from the views where it is visible to the views where it is occluded. MMA addresses this by letting each view selectively query the other views at the token level, so that spatially corresponding or semantically related patches can exchange information before reconstruction. This cross-view exchange is what enables the model to produce view-consistent anomaly maps and more reliable sample-level decisions.
}

Concretely, for each view $v$, MMA computes queries from $v$'s own tokens and aggregates key-value information from the remaining $V{-}1$ views, weighted by learned per-view importance scores.
A straightforward implementation would concatenate all $N$ patch tokens from all $V$ views into a single sequence of length $VN$ and apply standard softmax attention, but this yields $\mathcal{O}(V^2 N^2)$ complexity for the attention matrix alone, which quickly becomes prohibitive for high-resolution ViT features.
MMA sidesteps this bottleneck by leveraging the associativity of linear attention. Instead of materializing the full $VN \times VN$ attention matrix, it decomposes the cross-view computation into compact per-view statistics that can be accumulated in $\mathcal{O}(Vd^2)$ space (where $d{=}D/N_h$ is the per-head dimension defined below).

\noindent \textbf{Linear Attention Kernel.}
All operations in MMA are performed in a multi-head fashion with $N_h$ heads of dimension $d = D/N_h$.
Given the token sequence $\mathbf{Z}_v$ of view $v$ after layer normalization, we project it into queries, keys, and values through separate linear layers and reshape the result into $N_h$ by
\begin{equation}
\mathbf{Q}_v \leftarrow  \frac{\phi(\mathbf{Z}_v \mathbf{W}^Q)}{\sqrt{d}}, \quad
\mathbf{K}_v \leftarrow  \frac{\phi(\mathbf{Z}_v \mathbf{W}^K)}{\sqrt{d}}, \quad
\mathbf{V}_v \leftarrow \mathbf{Z}_v \mathbf{W}^V,
\label{eq:qkv}
\end{equation}
where $\phi(\cdot) = \text{ReLU}(\cdot)^2$ is a non-negative kernel map~\cite{zhang2024relu} applied to the queries and keys.
This square ReLU kernel ensures that all attention weights remain non-negative, analogous to the softmax function, while being much cheaper to evaluate. The parameters in~(\ref{eq:qkv}), \ie, $\mathbf{W}^Q, \mathbf{W}^K, \mathbf{W}^V \in \mathbb{R}^{D \times D}$, are learnable projection matrices.
Compared to alternative non-negative kernels such as plain ReLU or ELU~\cite{clevert2015fast}, the squared form provides stronger expressiveness by amplifying large activations and suppressing near-zero ones.

\noindent \textbf{View-Weighted Aggregation.}
In multi-view imagery, a given surface patch is typically visible in only a subset of views, so not all views contribute equally to each query view.
To consider this, MMA learns a view importance matrix $\mathbf{W}_\text{view} \in \mathbb{R}^{V \times V}$ whose rows are normalized via softmax (excluding the diagonal, \ie, self-attention):
\begin{equation}
\label{eq:view_weight}
w_{vj} = \text{softmax}_{j \neq v}(\mathbf{W}_\text{view}[v, :])\,,
\end{equation}
where $w_{vj}$ indicates the learned importance of view $j$ to view $v$.
Rather than concatenating all other views' tokens into a single large attention matrix, MMA accumulates compact key-value statistics from each view independently.
For each view $v$, the cross-view output is computed as:
\begin{equation}
\label{eq:mma}
\text{MMA}(\mathbf{Q}_v) = \frac{\mathbf{Q}_v \displaystyle\sum_{j \neq v} w_{vj}\, \mathbf{K}_j^\top \mathbf{V}_j}{\mathbf{Q}_v \displaystyle\sum_{j \neq v} w_{vj}\, \mathbf{K}_j^\top \mathbf{1} + \epsilon}\,,
\end{equation}
where $\mathbf{K}_j^\top \mathbf{V}_j \in \mathbb{R}^{d \times d}$ is the outer-product summary of view $j$ and $\mathbf{K}_j^\top \mathbf{1} \in \mathbb{R}^{d}$ is the normalization term.
The key insight is that these per-view statistics can be computed and accumulated \emph{before} multiplying with $\mathbf{Q}_v$, reducing the per-view complexity from $\mathcal{O}(N^2 d)$ in standard attention to $\mathcal{O}(Nd^2)$ (linear in the number of tokens $N$).
The attention output is then layer-normalized and projected back to $D$ dimensions: $\text{LN}(\text{MMA}(\mathbf{Q}_v))\,\mathbf{W}^O$, where $\mathbf{W}^O \in \mathbb{R}^{D \times D}$.

\input{tab_experiments_realiad}

\input{tab_experiments_manta}

\noindent \textbf{Token-wise Gating.}
Not every token benefits equally from cross-view fusion.
For instance, a patch depicting a featureless background carries little useful information for other views, while a patch near a defect boundary may gain significantly from complementary viewpoints.
A uniform residual connection would blend cross-view signals indiscriminately, potentially introducing noise from uninformative views.
To address this, we employ a token-wise gate that lets each token independently control its fusion ratio:
\begin{equation}
\label{eq:mma_gate}
\hat{\mathbf{Z}}_v = (1 - \mathbf{G}_v) \odot \mathbf{Z}_v + \mathbf{G}_v \odot \text{MMA}(\mathbf{Q}_v)\,,
\end{equation}
where $\odot$ is element-wise multiplication, $\mathbf{G}_v = \sigma(\text{MLP}(\text{MMA}(\mathbf{Q}_v))) \in [0,1]^{N \times D}$ and $\sigma$ denotes the sigmoid function.
The gate is predicted from the fused features themselves, allowing the network to learn which tokens contain informative cross-view evidence and should be integrated, and which should retain their original single-view representation.
A final layer normalization is applied to $\hat{\mathbf{Z}}_v$ before passing it to the subsequent stage.

In summary, MMA enriches each view's token representation with fine-grained evidence from other views through three cooperating mechanisms: a linear attention kernel for efficient cross-view matching, view-level importance weighting for selective aggregation, and token-wise gating for adaptive fusion.
Because the decoder is trained to reconstruct \emph{normal} features, tokens that have been enriched with cross-view context produce a larger reconstruction gap when an anomaly is present across views, strengthening both localization and detection.

%% file: tab_experiments_realiad.tex
\begin{table*}[t!]
\caption{
Average performance comparison on Real-IAD dataset. The best and second-best results are marked in \textbf{bold} and \ul{underlined}. All methods except MVAD are reproduced under the multi-view setting. 
}
\vspace{-1em}
\centering
\resizebox{.85\linewidth}{!}{
\renewcommand{\arraystretch}{1}
\begin{tabular}{@{}cccccccccccc@{}}
\toprule
Method & Venue & S-AUROC & S-AP & S-F1 & I-AUROC & I-AP & I-F1 & P-AUROC & P-AP & P-F1 & P-AUPRO \\
\midrule
RD4AD~\cite{deng2022anomaly} & CVPR'22 & 85.7 & 92.4 & 87.7 & 83.0 & 79.6 & 74.4 & 97.3 & 25.8 & 33.5 & 90.4 \\
UniAD~\cite{you2022unified} & NeurIPS'22 & 87.3 & 93.1 & 88.9 & 83.0 & 83.4 & 74.6 & 97.4 & 23.7 & 31.4 & 87.3 \\
RD++~\cite{tien2023revisiting} & CVPR'23 & 87.0 & 93.3 & 88.2 & 83.7 & 80.9 & 74.9 & 97.7 & 25.8 & 33.5 & 90.0 \\
SimpleNet~\cite{liu2023simplenet} & CVPR'23 & 61.2 & 77.9 & 82.1 & 82.6 & 79.2 & 74.1 & 76.3 & 3.4 & 6.9 & 42.5 \\
DeSTSeg~\cite{zhang2023destseg} & CVPR'23 & 89.0 & 94.6 & 89.2 & 82.3 & 79.2 & 73.2 & 94.6 & 37.9 & 41.7 & 40.6 \\
RealNet~\cite{zhang2024realnet} & CVPR'24 & 72.0 & 87.1 & 82.5 & 65.2 & 65.4 & 62.7 & 60.5 & 27.2 & 24.9 & 28.7 \\
ViTAD~\cite{zhang2025exploring} & CVIU'25 & 88.8 & 94.3 & 89.3 & 82.8 & 80.0 & 73.7 & 97.1 & 23.9 & 31.8 & 84.3 \\
MVAD~\cite{he2024learning} & TMM'25 & 90.2 & 95.3 & 90.1 & 86.6 & 84.8 & \ul{77.2} & \ul{97.9} & 30.3 & 36.8 & 91.2 \\
Dinomaly~\cite{guo2025dinomaly} & CVPR'25 & \ul{92.3} & \ul{96.2} & \textbf{91.7} & \ul{89.4} & \ul{86.8} & \textbf{80.3} & \textbf{98.9} & \ul{43.1} & \ul{47.3} & \ul{94.0} \\
\ourso & - & \textbf{92.4} & \textbf{96.3} & \ul{91.4} & \textbf{89.7} & \textbf{87.2} & \textbf{80.3} & \textbf{98.9} & \textbf{44.7} & \textbf{48.2} & \textbf{94.9} \\

\bottomrule
\end{tabular}
}
\vspace{-1em}
\label{tab:realiad_average}
\end{table*}

%% file: tab_experiments_manta.tex
\begin{table*}[t!]
\caption{
Average performance comparison on MANTA dataset. The best and second-best results are marked in \textbf{bold} and \ul{underlined}. All methods except MVAD are reproduced under the multi-view setting.}
\vspace{-1em}
\centering
\resizebox{.85\linewidth}{!}{
\renewcommand{\arraystretch}{1}
\begin{tabular}{@{}ccccccccccccc@{}}
\toprule
Method & Venue & S-AUROC & S-AP & S-F1 & I-AUROC & I-AP & I-F1 & P-AUROC & P-AP & P-F1 & P-AUPRO \\
\midrule
RD4AD~\cite{deng2022anomaly} & CVPR'22 & 78.3 & 63.7 & 68.3 & 80.5 & 59.5 & 62.4 & 91.8 & 29.1 & 34.7 & 78.8 \\
UniAD~\cite{you2022unified} & NeurIPS'22 & 74.9 & 59.7 & 65.2 & 82.7 & 64.3 & 65.1 & 90.7 & 26.1 & 32.4 & 79.6 \\
DeSTSeg~\cite{zhang2023destseg} & CVPR'23 & 62.4 & 49.3 & 56.5 & 62.3 & 37.8 & 45.5 & 54.8 & 12.0 & 9.0 & 23.4 \\
RealNet~\cite{zhang2024realnet} & CVPR'24 & 55.1 & 40.9 & 51.7 & 57.3 & 31.8 & 41.3 & 50.0 & 10.3 & 4.4 & 15.1 \\
ViTAD~\cite{zhang2025exploring} & CVIU'25 & 87.3 & 80.5 & 75.9 & 87.7 & 75.3 & 71.0 & 94.7 & 39.4 & 43.1 & 80.2 \\
MVAD~\cite{he2024learning} & TMM'25 & 85.0 & 76.5 & 73.8 & 85.2 & 69.9 & 67.9 & 93.7 & 32.7 & 38.5 & 80.7 \\
Dinomaly~\cite{guo2025dinomaly} & CVPR'25 & \ul{92.3} & \ul{87.9} & \ul{83.4} & \ul{91.3} & \ul{82.3} & \ul{77.7} & \ul{94.8} & \ul{44.8} & \ul{48.3} & \ul{84.2} \\
\ourso & - & \textbf{92.7} & \textbf{88.5} & \textbf{83.6} & \textbf{92.0} & \textbf{83.5} & \textbf{78.1} & \textbf{95.2} & \textbf{47.6} & \textbf{49.6} & \textbf{85.9} \\
\bottomrule
\end{tabular}
}
\vspace{-.5em}
\label{tab:manta_tiny_average}
\end{table*}

%% file: 4_experiments.tex
\section{Experiments}
\label{sec:experiments}

\subsection{Implementation Details}
We adopt a frozen DINOv2 ViT-Base/16~\cite{oquab2023dinov2} as the encoder backbone and extract intermediate features from the layers 2 through 9, which are averaged into a single token sequence per view.
The decoder consists of 8 Transformer blocks with 12 attention heads and an MLP ratio of 4.
A single-layer MLP bottleneck with a dropout rate of 0.4 connects the encoder features to the decoder.
For multi-view fusion, MMA employs linear attention with a ReLU$^2$ kernel and token-wise gating, while OGA uses cosine-similarity attention with a temperature of 0.7.
We train with StableAdamW~\cite{wortsman2023stable} using a learning rate of $2\!\times\!10^{-3}$, weight decay of $10^{-4}$, and a warm-cosine schedule that decays the learning rate to $2\!\times\!10^{-4}$ over 50K iterations after 100 warm-up steps.
The images are resized to $448\!\times\!448$ (crop $392\!\times\!392$) for Real-IAD and $280\!\times\!280$ (crop $266\!\times\!266$) for MANTA-Tiny.
The batch size is 4 for both datasets.

\subsection{Datasets}
We evaluate \ours on two multi-view AD benchmarks that differ in object scale and domain coverage.
\textbf{Real-IAD}~\cite{wang2024real} is a large-scale multi-view industrial AD benchmark covering 30 object classes with materials ranging from metal and plastic to wood and ceramics. It provides approximately 150K high-resolution images (99,721 normal, 51,329 abnormal) captured from five fixed viewpoints per object. Training uses only normal samples, while testing includes both normal and abnormal ones. Eight defect categories are annotated (pit, deformation, scratch, \etc), and defect areas span 0.1\% to 6.75\% of the image, posing a significant localization challenge.
\textbf{MANTA}~\cite{fan2024manta} targets tiny-object AD across 38 classes in five domains (mechanics, medicine, electronics, agriculture, and food), offering over 137.3K object instances with 8.6K abnormal samples annotated at the pixel level.
\SC{We use the official MANTA-Tiny split, which contains 800 instances per class (4,000 images per class over 5 views, 152,000 images in total). The full MANTA dataset contains images of varying sizes that cannot be evenly partitioned into five fixed views, whereas MANTA-Tiny provides uniformly sized 256$\times$256 images suitable for multi-view evaluation. Because \ours relies solely on visual information, we discard the textual annotations included in the original MANTA release.}

\input{tab_ablation}

\vspace{-1em}
\subsection{Metrics}
We adopt the evaluation protocol of~\cite{he2024learning} and report metrics at three granularities.
\textit{Sample-level} metrics (S-AUROC, S-AP, S-F1) judge whether an object instance is anomalous by combining evidence from all its views.
\textit{Image-level} metrics (I-AUROC, I-AP, I-F1) measure per-image detection accuracy independently of multi-view aggregation.
\textit{Pixel-level} metrics (P-AUROC, P-AP, P-F1, P-AUPRO) quantify how precisely the model localizes defective regions, where P-AUPRO (Per-Region-Overlap) penalizes large false-positive areas.
In ablation studies, we report four key indicators (S-AUROC, I-AUROC, P-AUROC, P-AUPRO) for brevity.

\subsection{Comparison with the State-of-the-Art}

We compare \ours with a range of AD methods, including unified models~\cite{you2022unified}, reconstruction-based approaches~\cite{deng2022anomaly, tien2023revisiting, guo2025dinomaly}, feature-embedding methods~\cite{liu2023simplenet, zhang2024realnet}, segmentation-based methods~\cite{zhang2023destseg}, and the multi-view method MVAD~\cite{he2024learning}. Since most existing methods are designed for single-view evaluation, we reproduce them under the multi-view setting for a fair comparison. MVAD is evaluated directly as a multi-view method.

\input{tab_ab_crossview}

\input{tab_ab_crossview_performance}

\noindent \textbf{Results on Real-IAD.}
Table~\ref{tab:realiad_average} summarizes the results on Real-IAD. \ours performs the best on 9 out of 10 individual metrics, producing sample-level scores of 92.4/96.3/91.4, image-level scores of 89.7/87.2/80.3, and pixel-level scores of 98.9/44.7/48.2/94.9. Compared with Dinomaly, the strongest single-view baseline, the most notable gains appear at the pixel level, where MMA and OGA provide spatial cues from complementary viewpoints that help the model produce tighter anomaly maps: P-AP improves by +1.6 (43.1$\to$44.7) and P-F1 by +0.9 (47.3$\to$48.2), indicating sharper anomaly localization, while P-AUPRO rises by +0.9 (94.0$\to$94.9), confirming that region-level coverage improves as well. Compared with MVAD, the only other multi-view method, \ours achieves substantially stronger results across all metrics. The localization gap is particularly striking: +14.4 on P-AP and +11.4 on P-F1, demonstrating that our global-local fusion design produces more precise anomaly maps than its cross-view attention mechanism.

\input{tab_ab_OGA}

\noindent \textbf{Results on MANTA-Tiny.}
Table~\ref{tab:manta_tiny_average} reports results on MANTA-Tiny. \ours performs the best on all 10 individual metrics with sample-level scores of 92.7/88.5/83.6, image-level scores of 92.0/83.5/78.1, and pixel-level scores of 95.2/47.6/49.6/85.9. The pixel-level gains over Dinomaly are especially noteworthy: P-AP improves by +2.8 (44.8$\to$47.6) and P-F1 by +1.3 (48.3$\to$49.6), representing the largest improvements among all metric groups. Cross-view fusion is particularly beneficial for tiny-object localization, where single-view features may be ambiguous due to the limited spatial extent of defects; complementary viewpoints supply the additional spatial context needed to resolve such ambiguities. P-AUPRO also rises by +1.7 (84.2$\to$85.9), confirming consistent region-level improvement. Overall, \ours consistently outperforms all compared methods on both datasets, validating the benefit of incorporating multi-view information through the proposed MMA and OGA modules.

\input{fig_qualitative}

\noindent \textbf{Qualitative Results.}
Figure~\ref{fig:qualitative} shows representative anomaly maps from both datasets. Consistent with the quantitative gains, \ours has noticeably fewer false positives and cleaner anomaly maps that focus activation on real defective regions. On Real-IAD \textit{Button Battery} class, the defect occupies only a small region in one view, yet \ours localizes it with a focused hot spot, while MVAD and Dinomaly spread activations across the entire surface. For \textit{PCB}, whose complex on-board components easily mislead other methods into diffuse responses, \ours suppresses background noise and assigns distinctly higher scores at ground-truth locations, reflecting the cross-view stability from OGA and MMA. 
On MANTA-Tiny \textit{Gear} and \textit{Flat Nut} classes, \ours yields the most saturated predictions aligned with the ground truth, with fewer missed detections and fewer false alarms, even without observing all defect types during training.

\subsection{Ablation Study}
\noindent \textbf{Component Analysis.}
Table~\ref{tab:ablation} reports the contribution of each proposed module. MMA and OGA address different aspects of the multi-view AD problem. Adding MMA alone primarily benefits sample-level detection, with S-AUROC improving by +2.2 on Real-IAD, as cross-view token fusion provides direct evidence for object-level decisions. However, MMA alone slightly lowers P-AP. 
Because MMA operates at the patch level, small anomalous regions can have their signal diluted by fused cross-view features.
OGA tackles this from a different angle. By aggregating class tokens into a global object representation, it widens the gap between normal and abnormal feature distributions, which in turn sharpens pixel-level localization. 
Adding OGA yields substantial P-AP gains: on Real-IAD, P-AP rises from 42.7 to 44.7 (+2.0) and P-AUPRO from 94.4 to 94.9 (+0.5); on MANTA-Tiny, P-AP jumps from 44.6 to 47.6 (+3.0) and P-AUPRO from 85.1 to 85.9 (+0.8). Overall, the full model improves P-AUPRO by +0.9 on Real-IAD and +1.7 on MANTA-Tiny over the baseline without any multi-view module.

\input{tab_missingview}

\noindent \textbf{Cross-View Fusion Mechanism.}
We compare three alternatives for the cross-view attention in MMA: our linear attention variant, deformable attention, and the MVAS block from MVAD~\cite{he2024learning}. Table~\ref{tab:ablation_cross_view} shows that linear attention is the parameter-efficient (2.66M) and the fastest (12.04\,ms per forward pass). MVAS requires 3.86M parameters, 47.51\,ms, and 9,276\,MB during training (over 7$\times$ the memory of linear attention, 1,312.9\,MB); deformable attention is more compact (2.37M) but still costs 3,091.1\,MB. Table~\ref{tab:ablation_cross_view_performance} shows that the two mechanisms achieve comparable pixel-level detection of 98.9/42.7/47.0/94.4 and 98.9/43.3/46.7/94.4, respectively. Given the large efficiency advantage at nearly identical accuracy, linear attention is the preferred choice for MMA.

\noindent \textbf{Information Restriction in OGA.}
In standard multi-view learning, richer cross-view information exchange is generally beneficial. Multi-view AD, however, presents a counter-intuitive challenge that we term as \emph{cross-view information leakage}. 
Reconstruction-based methods rely on the gap between normal and anomalous reconstruction errors. When the global attention module passes too many normal cues across views, the decoder can faithfully reconstruct even anomalous regions, collapsing this gap.
As shown in Table~\ref{tab:ablation_oga}, we introduce four complementary restrictions into OGA.
The most impactful is \emph{token replacement}, which directly replaces the original patch tokens with the cross-attention output instead of adding a residual. All subsequent variants build on token replacement.
Second, we replace softmax with \emph{sigmoid} gating, which permits sparse, selective activation rather than forcing a normalized distribution that spreads information uniformly across views. Third, the \emph{NoGlobal} strategy withholds the aggregated global object token from per-view CLS tokens, preventing multi-view information from leaking into the per-view classification signal. Fourth, a sharpened \emph{temperature} ($\tau{=}0.7$) concentrates the attention distribution to further limit diffuse information spread. Table~\ref{tab:ablation_oga} shows that these restrictions yield consistent improvements on MANTA-Tiny.
Notably, sample-level metrics remain stable or even improve alongside the pixel-level gains, confirming that the restrictions successfully prevent information leakage without sacrificing detection capability. This finding suggests that information restriction is a fundamental design principle for multi-view AD, distinct from the ``more information is better'' paradigm that prevails in general multi-view representation learning.

\noindent \textbf{Robustness to Missing Views.} Table~\ref{tab:missingview} shows that \ours remains largely stable even with fewer views: P-AUROC stays nearly unchanged since each remaining view still provides its own local evidence for localization, and sample-level S-AUROC stays above 90 even after removing three of the five views.

%% file: tab_ablation.tex
\begin{table*}[t!]
\caption{
\SC{
Ablation study on the contribution of each proposed module. MMA primarily improves sample-level detection, while OGA further boosts pixel-level localization. The best results per dataset are marked in \textbf{bold}.
}
}
\vspace{-1em}
\centering
\resizebox{.88\linewidth}{!}{
\renewcommand{\arraystretch}{1}
\begin{tabular}{@{}ccccccccccccc@{}}
\toprule
Dataset & MMA & OGA & S-AUROC & S-AP & S-F1 & I-AUROC & I-AP & I-F1 & P-AUROC & P-AP & P-F1 & P-AUPRO \\
\midrule

\multirow{3}{*}{Real-IAD} & \ccross & \ccross & 90.3 & 96.2 & \textbf{91.7} & 89.4 & 86.8 & 80.3 & \textbf{98.9} & 43.1 & 47.3 & 94.0 \\
& \ccheck & \ccross & \textbf{92.5} & \textbf{96.3} & 91.6 & \textbf{89.8} & \textbf{87.2} & \textbf{80.8} & \textbf{98.9} & 42.7 & 47.0 & 94.4 \\
& \ccheck & \ccheck & 92.4 & \textbf{96.3} & 91.4 & 89.7 & \textbf{87.2} & 80.3 & \textbf{98.9} & \textbf{44.7} & \textbf{48.2} & \textbf{94.9}  \\
\midrule

\multirow{3}{*}{MANTA} & \ccross & \ccross & 92.3 & 87.9 & 83.4 & 91.3 & 82.3 & 77.7 & 94.8 & 44.8 & 48.3 & 84.2 \\
& \ccheck & \ccross & \textbf{92.8} & 88.3 & 83.4 & 91.8 & 82.8 & 77.9 & 94.9 & 44.6 & 47.6 & 85.1 \\
& \ccheck & \ccheck & 92.7 & \textbf{88.5} & \textbf{83.6} & \textbf{92.0} & \textbf{83.5} & \textbf{78.1} & \textbf{95.2} & \textbf{47.6} & \textbf{49.6} & \textbf{85.9} \\
\bottomrule

\end{tabular}
}
\vspace{-1em}
\label{tab:ablation}
\end{table*}

%% file: tab_ab_crossview.tex
\begin{table}[!t]
\caption{\SC{Efficiency comparison of cross-view fusion alternatives in the MMA module.}}
\vspace{-1em}
\setlength{\tabcolsep}{4pt}
\resizebox{\linewidth}{!}
{
\begin{tabular}{@{}c@{}cccc@{}}
\toprule

Approach & Params (M) & Fwd (MB) & Fwd+Bwd (MB) & Time (ms) \\
\midrule
Deformable Attention & \textbf{2.37} & 2208.1 & 3091.1 & 17.38 \\
MVAS Block~\cite{he2024learning} & 3.86 & 1318.1 & 9276.0 & 47.51 \\
Linear Attention (ours) & 2.66 & \textbf{528.5} & \textbf{1312.9} & \textbf{12.04} \\
\bottomrule
\end{tabular}
}
\vspace{-1em}
\label{tab:ablation_cross_view}
\end{table}

%% file: tab_ab_crossview_performance.tex
\begin{table}[!t]
\caption{\SC{Accuracy comparison of cross-view fusion alternatives in the MMA module.}}
\vspace{-1em}
\setlength{\tabcolsep}{4pt}
\resizebox{\linewidth}{!}
{
\begin{tabular}{@{}c@{}c@{\hspace{2mm}}c@{\hspace{2mm}}c@{\hspace{2mm}}c@{\hspace{2mm}}c@{\hspace{2mm}}c@{}}
\toprule

Approach & S-AUROC & I-AUROC & P-AUROC & P-AP & P-F1 & P-AUPRO \\
\midrule
Deformable Attention & 92.4 & 89.7 & \textbf{98.9} & \textbf{43.3} & 46.7 & \textbf{94.4} \\
Linear Attention (ours) & \textbf{92.5} & \textbf{89.8} & \textbf{98.9} & 42.7 & \textbf{47.0} & \textbf{94.4} \\
\bottomrule
\end{tabular}
}
\vspace{-1em}
\label{tab:ablation_cross_view_performance}
\end{table}

%% file: tab_ab_OGA.tex
\begin{table}[!t]
\caption{
Ablation study on OGA information-restriction mechanisms. All variants include MMA and OGA. TR.: token replacement; SG.: sigmoid gating; NoG.: no global token feedback to per-view CLS tokens; Temp.: sharpened temperature. The best results are marked in \textbf{bold}.
}
\vspace{-1em}
\centering
\resizebox{\linewidth}{!}{
\renewcommand{\arraystretch}{1}
\begin{tabular}{@{}c@{\hspace{2mm}}c@{\hspace{2mm}}c@{\hspace{2mm}}c@{\hspace{2mm}}c@{\hspace{2mm}}c@{\hspace{2mm}}c@{\hspace{2mm}}c@{\hspace{2mm}}c@{\hspace{2mm}}c@{}}
\toprule
TR. & SG. & NoG. & Temp. & S-AUROC & I-AUROC & P-AUROC & P-AP & P-F1 & P-AUPRO \\
\midrule
\ccross & \ccross & \ccross & \ccross & 92.2 & 91.3 & 94.7 & 43.7 & 46.9 & 84.0 \\
\ccheck & \ccross & \ccross & \ccross & 92.4 & 91.7 & 95.0 & 47.0 & 49.4 & 85.4 \\
\ccheck & \ccheck & \ccross & \ccross & 92.3 & 91.8 & 95.1 & 46.8 & 49.3 & 85.4 \\
\ccheck & \ccross & \ccheck & \ccross & 92.4 & 91.7 & 95.1 & 47.7 & 50.0 & 86.0 \\
\ccheck & \ccheck & \ccheck & \ccross & 92.6 & 91.9 & \textbf{95.3} & 47.7 & 49.8 & 85.7 \\
\ccheck & \ccheck & \ccross & \ccheck & 92.4 & 91.8 & 95.2 & \textbf{47.8} & \textbf{50.1} & \textbf{85.9} \\
\ccheck & \ccheck & \ccheck & \ccheck & \textbf{92.7} & \textbf{92.0} & 95.2 & 47.6 & 49.6 & \textbf{85.9} \\
\bottomrule
\end{tabular}
}
\vspace{-1.5em}
\label{tab:ablation_oga}
\end{table}

%% file: fig_qualitative.tex
\begin{figure*}[ht!]
    \centering
    \includegraphics[width=.97\textwidth]{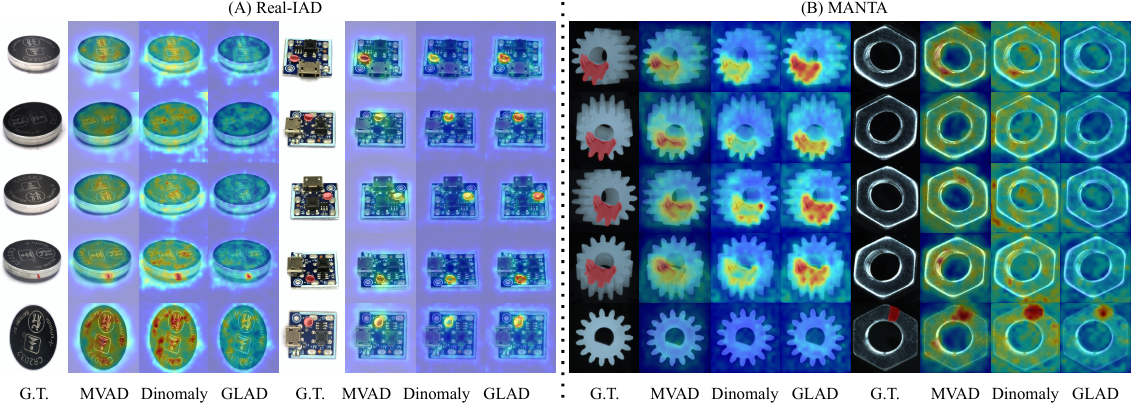}
    \Description[<short description>]{<long description>} 
    \vspace{-1em}
    \caption{
    \SC{Qualitative comparison of anomaly maps on (A) Real-IAD and (B) MANTA. We show the ground-truth (G.T.) annotation and the anomaly maps produced by MVAD, Dinomaly, and \ourso. Warmer colors indicate higher anomaly scores.}
    }
    \vspace{-1em}
    \label{fig:qualitative}
\end{figure*}

%% file: tab_missingview.tex
\begin{table}[t!]
\caption{\textbf{Missing-view experiment.} We mask out C2, C3, and C4 and renormalize view-importance weights over the remaining views without retraining.}
\vspace{-1em}
\centering
\renewcommand{\arraystretch}{1.1}
\resizebox{\columnwidth}{!}{%
\begin{tabular}{l cc cc cccc}
\toprule
Setting & S-AUROC & S-AP & I-AUROC & I-AP & P-AUROC & P-AP & P-F1 & P-AUPRO \\
\midrule
Baseline (all 5)   & 92.37 & 96.28 & 89.73 & 87.25 & 98.93 & 44.67 & 48.26 & 94.89 \\
w/o C2        & 91.13 & 94.84 & 89.44 & 87.00 & 98.92 & 44.74 & 48.38 & 94.80 \\
w/o C2,C3     & 90.57 & 93.11 & 89.10 & 86.50 & 98.91 & 44.55 & 48.18 & 94.78 \\
w/o C2,C3,C4  & 90.12 & 89.81 & 88.99 & 85.66 & 98.83 & 44.02 & 47.84 & 94.58 \\
\bottomrule
\end{tabular}%
}
\vspace{-1em}
\label{tab:missingview}
\end{table}

%% file: 5_conclusion.tex
\section{Conclusion}
\label{sec:conclusion}
We presented \ours, a global-local multi-view anomaly detection framework that integrates cross-view learning through two modules. 
\SC{
Through this work, we identified cross-view information leakage as a fundamental challenge in reconstruction-based multi-view anomaly detection, where naive fusion collapses the reconstruction gap by passing normal cues across views.
OGA addresses this through deliberate information restriction, preserving the reconstruction gap essential for detection, while MMA enables efficient token-level cross-view fusion with linear complexity.
Our experiments on Real-IAD and MANTA confirm that \ours outperforms state-of-the-art methods. We believe that the principle of information restriction offers a useful perspective for future work on multi-view reasoning in reconstruction-based settings beyond anomaly detection.}